\documentclass{bmvc2k}
\usepackage{booktabs}
\usepackage{multirow}
\usepackage{subcaption}
\usepackage{graphicx}
\usepackage{amsmath, amssymb, amsfonts}

\title{Hierarchical Image-Guided 3D Point Cloud Segmentation in Industrial Scenes via Multi-View Bayesian Fusion}

\addauthor{Yu Zhu}{yu.zhu.s3@dc.tohoku.ac.jp}{1}
\addauthor{Naoya Chiba}{chiba@chiba.net}{2}
\addauthor{Koichi Hashimoto}{koichi.hashimoto.a8@tohoku.ac.jp}{1}

\addinstitution{
  Department of System Information Sciences, Graduate School of Information Sciences \\
  Tohoku University\\
  Sendai, Japan
}

\addinstitution{
 D3 Center\\
 The University of Osaka \\
 Osaka, Japan
} 

\runninghead{ZHU ET AL,}{Hierarchical 3D Segmentation in Industrial Scenes}

\begin{document}

\maketitle

\begin{abstract}
    Reliable 3D segmentation is critical for understanding complex scenes with dense layouts and multi-scale objects, as commonly seen in industrial environments. In such scenarios, heavy occlusion weakens geometric boundaries between objects, and large differences in object scale will cause end-to-end models fail to capture both coarse and fine details accurately. Existing 3D point-based methods require costly annotations, while image-guided methods often suffer from semantic inconsistencies across views. To address these challenges, we propose a hierarchical image-guided 3D segmentation framework that progressively refines segmentation from instance-level to part-level.  Instance segmentation involves rendering a top-view image and projecting SAM-generated masks prompted by YOLO-World back onto the 3D point cloud. Part-level segmentation is subsequently performed by rendering multi-view images of each instance obtained from the previous stage and applying the same 2D segmentation and back-projection process at each view, followed by Bayesian updating fusion to ensure semantic consistency across views. Experiments on real-world factory data demonstrate that our method effectively handles occlusion and structural complexity, achieving consistently high per-class mIoU scores. Additional evaluations on public dataset confirm the generalization ability of our framework, highlighting its robustness, annotation efficiency, and adaptability to diverse 3D environments.
\end{abstract}
\section{Introduction}
\label{sec:intro}

In recent years, 3D scene understanding has become critical for perception tasks such as robotic manipulation and digital twin construction\cite{3Dseg_survey, 3Dseg_survey2}. These applications require precise recognition of both object instances and their structure, where fine-grained 3D segmentation plays a central role by inferring category labels for individual points in the scene to enable detailed geometric and semantic interpretation. Among various real-world scenarios, industrial environments are particularly challenging for 3D segmentation due to tightly connected components and complex spatial layouts. These characteristics often lead to visual clutter and ambiguous boundaries, making it difficult to separate individual parts and perform accurate segmentation.

Despite recent progress in 3D point cloud segmentation, existing methods still face significant challenges in industrial environments. Point-based approaches \cite{3D_POINTNET++,3D_GAISSIAN_GROUPING,3D_UNSCENE3D,3D_Memory-based,3Dpointcnn} require extensive manual annotations and often generalize poorly to large-scale, cluttered scenes. In contrast, image-based methods \cite{MULTISAM_OVIR3D,MULTISAM_SAI3D,MULTISAM_SAM3D}, powered by foundation models such as SAM\cite{SAM1,SAM2}, YOLO-World\cite{YoloWorld}, and related models, provide stronger representations and better transferability. However, per-view occlusion and feature inconsistency often lead to semantic inconsistencies, reducing the reliability of 3D fusion. Moreover, the lack of datasets tailored for industrial scenarios and the limitations of single end-to-end models make it difficult to segment both large objects and their fine-grained components, especially under multi-scale conditions.

To address these issues, we propose a hierarchical image-guided 3D segmentation framework consisting of two stages: instance-level and part-level segmentation. We fine-tune YOLO-World on a small set of industrial samples and use its predictions to prompt SAM. While SAM generates geometry-awared masks, it lacks semantic discrimination, which is compensated by YOLO-World’s categories prompts. Both stages follow a detect-then-segment strategy. In the first stage, 2D instance masks are generated from a top-view rendering image, allowing efficient extraction of each object's 3D points from a global view. In the second stage, we render multi-view images of each object to capture component details. The resulting part-level 2D masks are back-projected to the 3D point cloud and fused using Bayesian updating\cite{Bayes} to enforce semantic consistency across views.

Compared to end-to-end segmentation networks, this design reflects the semantic hierarchy of real-world scenarios while improving interpretability and flexibility through a modular and staged architecture. Our work leverages pre-trained 2D models with minimal supervision: the detection model is fine-tuned on a small custom dataset. This allows rapid adaptation to new scenario and tasks and significantly lowers annotation and training costs. 
The main contributions are as follows:
\begin{itemize}
\item We propose a hierarchical image-guided framework that decomposes 3D segmentation into instance-to-part and detection-to-segmentation stages, improving accuracy and interpretability over single-model approaches.
\item We introduce a Bayesian updating fusion mechanism that addresses cross-view inconsistency and significantly enhances part-level segmentation accuracy under occlusion and viewpoint variation.
\item We provide a modular pipeline based on vision foundation models that supports accurate and interpretable 3D segmentation with lower annotation and training cost, enabling fast adaptation to various complex 3D environments.
\end{itemize}

\begin{figure}[!htbp]
  \centering
  \includegraphics[width=0.9\linewidth]{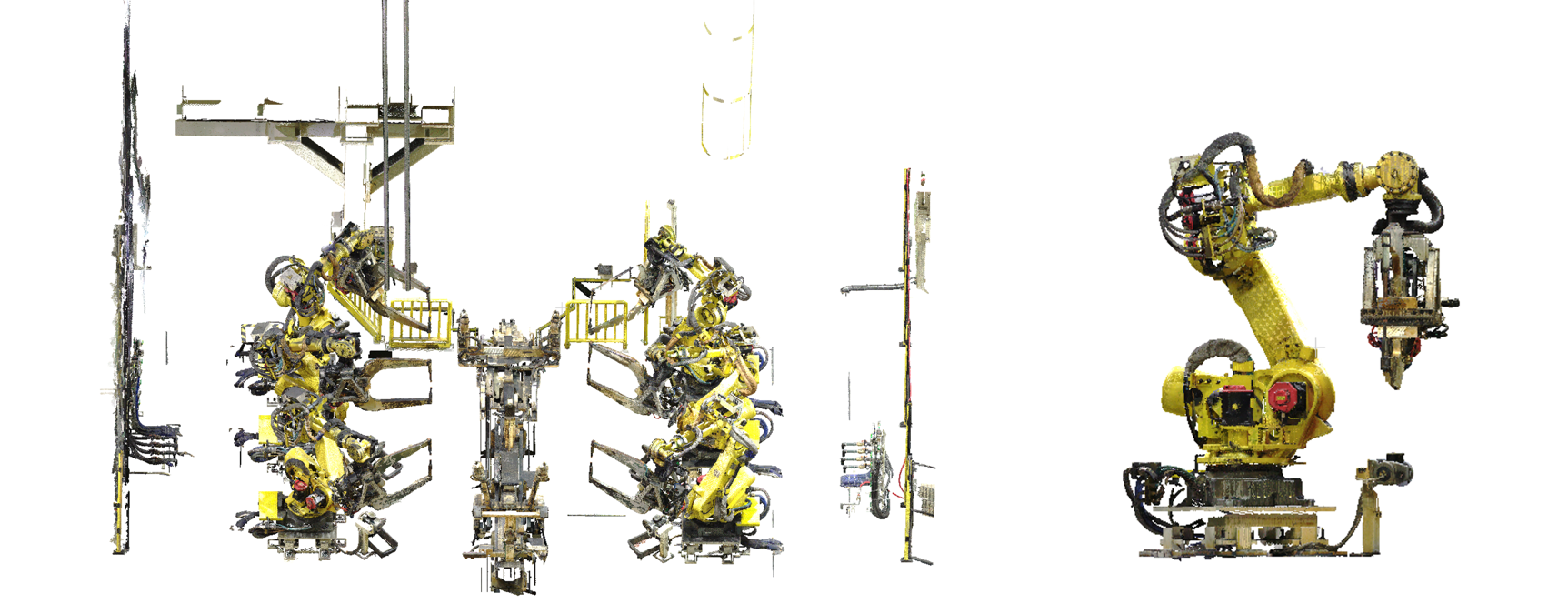}
  \vspace{0.4em}
  \caption{(Left) Entire manufacturing factory. (Right) Close-up view of RA equipment.}
  \label{fig:factory_minipage}
\end{figure}

\section{Related Works}
\label{sec:relworks}

\textbf{Direct 3D Point Cloud Segmentation} Traditional approaches use grouping methods\cite{Grouping,Grouping_HierarchicalPointGrouping,Grouping_PointGroup,Grouping_SGPN,Grouping_softgroup}, but these approaches rely on geometric proximity and handcrafted heuristics, limiting their scalability to complex scenes. Deep neural network-based segmentation has been widely used in recent years, like\cite{3D_POINTNET++, 3D_GAISSIAN_GROUPING, 3D_UNSCENE3D, 3D_Memory-based, 3Dpointcnn}, which learn features from unordered points to predict semantic labels directly. However, due to the lack of explicit topology and adjacency—unlike the regular structure in images, these models struggle to capture stable local geometry. As a result, they require dense point-level annotations and often perform poorly in large-scale, cluttered scenes. Moreover, existing 3D datasets such as\cite{DATASET1,DATASET2,DATASET_SCANNET,DATASET_3RSCAN,DATASET_partnet,DATASET_3DSUNRGBD,DATASET_SCANNET++} with annotations are small, as annotating detailed 3D models is labor-intensive. In particular, to the best of our knowledge, no public dataset currently supports part-level segmentation in industrial environments.

\textbf{Image-Guided 3D Segmentation via 2D Supervision} To overcome the limitations of 3D supervision, some methods\cite{MULTISAM_OVIR3D, MULTISAM_SAI3D, MULTISAM_SAM3D} project 3D point clouds onto 2D images and use the resulting masks to transfer 2D semantics into 3D space. Vision foundation models such as SAM\cite{SAM1,SAM2}, YOLO-World\cite{YoloWorld}, and GLIP\cite{GLIP,GLIPv2} benefit from large-scale 2D datasets and weakly supervised training (e.g., with bounding boxes), enabling high-quality segmentation at low annotation cost. However, SAM does not provide category labels, while such as YOLO-World and GLIP offer only class predictions without pixel-level masks, making them difficult to apply to 3D segmentation. Rendering point clouds into 2D images introduces occlusions and viewpoint-dependent variations, often causing inconsistent boundaries and semantics across views. Without a robust fusion strategy, current methods cannot ensure inconsistency 3D segmentation. Moreover, most foundation models are trained on general categories and cannot directly adapt to factory scenes. Existing 2D image datasets such as \cite{DATASET_MVIMGNET,DATASET_sun2020scalability,DATASET_nuScenes} also lack coverage of industrial environments.


\section{Proposed Method for 3D Point Cloud Segmentation of Manufacturing Scenes }
\label{sec:method}
\begin{figure}[!htbp]
  \centering
  \includegraphics[width=\textwidth]{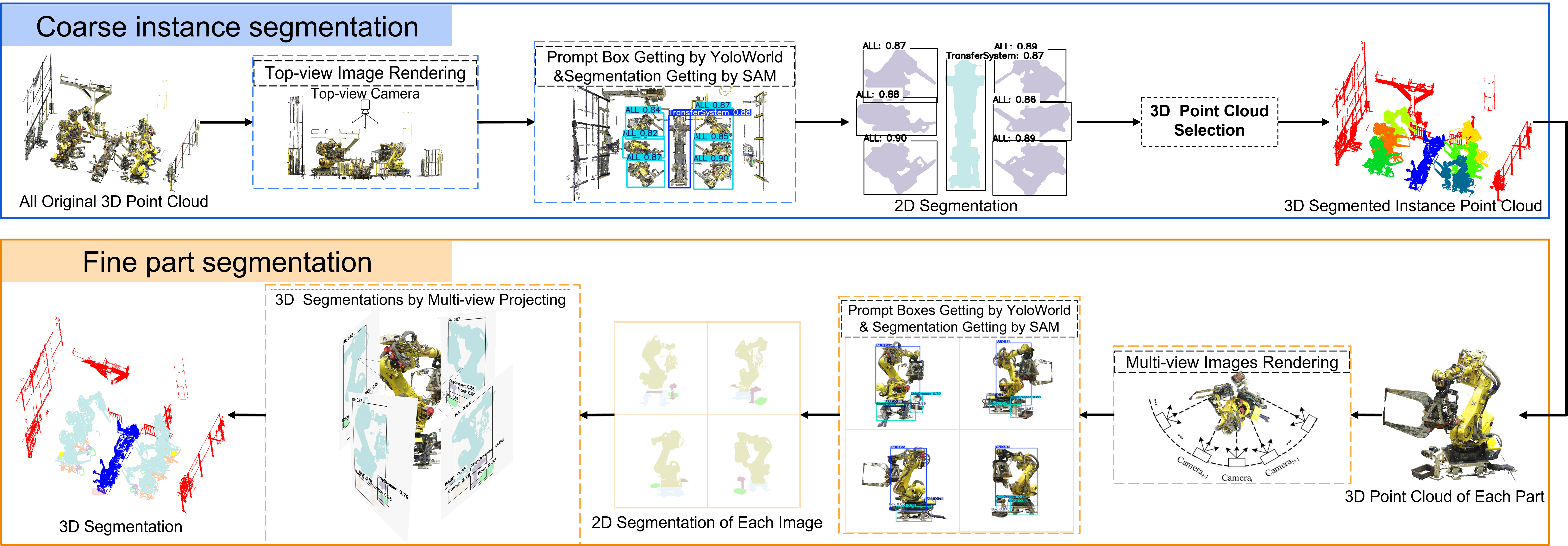}
  \vspace{0.4em}
  \caption{The pipeline of the proposed hierarchical 3D segmentation system. The framework consists of two main stages: (1) coarse instance-level segmentation; (2) fine-grained part-level segmentation. }
  \label{fig:Overview_of_System}
\end{figure}

As discussed in Section~\ref{sec:relworks}, existing methods are unsuitable for part-level segmentation in cluttered, hierarchical environments due to their reliance on costly 3D annotations or inconsistent multi-view fusion. Moreover, few frameworks are adaptable to real manufacturing scenes. To address this, we propose a two-stage hierarchical segmentation framework guided by 2D predictions, progressively refining scene understanding from coarse instances to fine-grained parts (Fig.~\ref{fig:Overview_of_System}). The proposed framework begins with adaptive rendering, which projects 3D point clouds into 2D images using scale-aware parameters based on object size and point density for each stage. In each stage, YOLO-World detects objects and provides prompts to the Segment Anything Model (SAM) to generate 2D masks, which are then back-projected into 3D. Instance-level segmentation uses top-view projection to efficiently label large objects, while part-level segmentation fuses multi-view masks via Bayesian updating to resolve inconsistencies and improve label stability under occlusion and viewpoint variation.

\subsection{Adaptive 2D Image Rendering and Segmentation}
In real manufacturing environments, 3D point clouds—whether captured from real scans or generated synthetically—often exhibit significant variation in object scale and sampling density. Fixed rendering parameters struggle to handle this diversity, resulting in over-cropped large objects or under-resolved small parts, both of which degrade segmentation accuracy.

To support high-quality segmentation across such variations, we propose a scale-adaptive rendering strategy that computes the point radius based on object geometry and density. The radius \( r \) is defined as:
\begin{equation}
    \text{r} = r_{\text{px}}\left( \frac{s}{I\cdot \rho} \right) 
\end{equation}
where \( r_{\text{px}} \) is the desired point size in pixels (typically 2–4), \( I \) is the image resolution, \( s \) is the maximum bounding box extent, and \( \rho \) denotes point cloud density in cubic centimeter.

The rendered 2D images are segmented using a two-stage method combining YOLO-World and SAM.  YOLO-World detects object instances or parts and generates class-aware bounding boxes, which serve as prompts for SAM to produce high-quality pixel-wise masks.  This prompt-based design enables SAM to generate accurate segmentations even under clutter, occlusion, and scale variation.  Unlike end-to-end models that directly predict masks from raw inputs, our detection-driven framework separates object localization from mask generation, improving flexibility, interpretability, and task transferability.



\subsection{3D Point Cloud Segmentation via 2D Mask Projection}

After obtaining accurate 2D masks at each stage, we transfer their geometric and semantic priors into 3D space. To address the differing goals of the two stages, we design separate projection methods for instance-level and part-level segmentation. 

\textbf{Instance-Level Segmentation from Top View}  
In this stage, where the point cloud is large and densely distributed, we adopt a lightweight and efficient strategy based on 2D mask projection. Given the input 3D point cloud \( \mathcal{P}\) and a set of 2D instance masks \( \{ \mathcal{M}_i^{top} \}_{i=1}^{k} \), where each \( \mathcal{M}_i^{top} \) denotes the 2D mask of instance \( i \), we project each point \( X \in \mathcal{P} \) onto the top-view image plane using known camera parameters. A point is assigned to instance \( i \) if its projection falls within the corresponding mask, as defined by:
\begin{equation}
\mathcal{\mathcal{P}}_i = \left\{ X \in \mathcal{P} \mid \pi(R_{top}X+t_{top}) \in \mathcal{M}_i^{top} \right\}
\end{equation}
where $\pi(\cdot)$ denotes the pinhole projection with intrinsics $K$, and 
$R_{(\cdot)} \in \mathit{SO}(3),\ t_{(\cdot)} \in \mathit{R}^3$.
 are the rotation and 
translation that transform points from the world frame to the corresponding camera frame. This projection-based filtering enables fast and coarse instance segmentation without relying on complex 3D computation, making it well-suited for large-scale and dense point clouds.

\textbf{Part-Level Segmentation via Multi-view Back-Projection}  
Unlike instance-level segmentation, part-level segmentation requires a more detailed understanding of internal geometric structures. To capture such fine-grained information, we sample \( n \) viewpoints \( \Theta = \{ \theta_1, \dots, \theta_n \} \) around each instance point cloud \( \mathcal{P}_i \). At each view \( \theta \in \Theta \), we obtain part-level 2D masks \( \{ \mathcal{M}_j^{\theta} \}_{j=1}^{k} \) using YOLO-World + SAM. Each pixel \( (u, v) \in \mathcal{M}_j^{\theta} \) is back-projected into a query point via:
\begin{equation}
  X_{\text{query}}^\theta = R_\theta^{-1} \left( \pi^{-1}(u, v, d_\theta(u,v)) - t_\theta \right)  
\end{equation}

We apply a depth-guided KD-tree\cite{KDTREE} search on \( \mathcal{P}_i \) to match each query point with its nearest neighbor. A raw point \( X \in \mathcal{P}_i \) is matched if \( \left\| X_{\text{query}}^\theta - X \right\|_2 < \varepsilon \), and its depth from the viewpoint is consistent with the rendered value.All pixel-wise semantic observations that contribute to point \( X \) are denoted as \( \mathcal{X} = \{ x^{\theta_1}, \dots, x^{\theta_n} \} \).


\subsection{Multi-view Mask Consistency with Bayesian Updating Fusion}
Due to geometric differences across viewpoints, semantic observations from back-projected masks may become inconsistent for the same 3D point. To address this, we estimate a consistent class distribution for each point \( X \) by recursively fusing its multi-view observations \( \{ x^{\theta_1}, \dots, x^{\theta_n} \} \) through Bayesian updating\cite{Bayes}. The class variable at view \( \theta \) is denoted as \( c_\theta \). Each observation \( x^\theta \) is modeled as a soft class distribution: the top-1 confidence from YOLO-World is assigned to the predicted class, while the remaining probability is uniformly distributed across the other classes. The posterior is updated at each view by combining the current observation with the prior from previous views:


\begin{equation}
p_\theta(c_\theta \mid \mathcal{X}^{{\theta}_0:{\theta}_n}) =
\frac{\alpha_\theta}{Z_\theta} \cdot p(x^\theta \mid c_\theta) \cdot p_{\theta_{n-1}}(c_\theta \mid \mathcal{X}^{{\theta}_0:{\theta}_{n-1}})
\end{equation}

Which \( Z_\theta \) is the normalization constant ensuring \( \sum_{c_\theta} p_\theta(c_\theta \mid \mathcal{X}^{\theta_0:\theta_n}) = 1 \). To further improve robustness under occlusion and noise, we introduce a geometry-aware confidence score \( \alpha_\theta \in [0, 1] \) for each view, computed as:

\begin{equation}
\alpha_\theta = \frac{1}{3}(A + N + B)
\end{equation}

\newpage\noindent
where:
\begin{itemize}
  \item \( A = \frac{\text{area}( \mathcal{M}_j^{\theta})}{H \times W} \) is the normalized mask area, clipped to \([0.001, 0.5]\) to suppress degenerate tiny/overly-large masks;
  \item \( N \in \{0.5, 1.0\} \) is a binary score indicating whether the number of projected 3D points within the mask exceeds a predefined threshold (e.g., 100) to suppress unreliable observations caused by sparse points due to noise;
  \item \( B \in \{0.5, 1.0\} \) captures boundary complexity to down-weight
clearly under-segmented or over-fragmented masks;
\end{itemize}
Using the same setting $A/N/B$ configuration on both industrial scenes and
PartNet\cite{DATASET_partnet} dataset to prove that our view-confidence is generic and insensitive to dataset specifics.

For each point \( X \in \mathcal{P}_i \), the final set of labeled points is constructed by selecting  the most probable class label if its confidence exceeds a threshold \( \tau \):

\begin{equation}
\mathcal{P}_{\text{part}_m} = \bigcup_{X \in \mathcal{P}_i} \left\{ (X, \arg\max_c p(c)) \mid \max_c p(c) > \tau \right\}
\end{equation}
Through Bayesian updating fusion, each part within an instance is segmented with semantic and geometric consistency. However, residual noise may still exist due to imperfect projections or local ambiguities. To address this, we apply a DBSCAN\cite{DBSCAN} clustering step to remove outliers and further improve the accuracy of part-level segmentation across the entire scene.

\section{3D Point Cloud Segmentation Experimental Details}
\label{sec:experiments}
We evaluate the performance of each module and the overall system. Specifically, we conducted the following three experiments, including the evaluation of the 2D recognition and segmentation model, the evaluation of 3D projection fusion, and the performance of the overall algorithm on a general open-source dataset and real industry scenarios. All experiments were conducted on a machine equipped with an Intel i9-13900HK (2.60 GHz) CPU, NVIDIA RTX 4090 Laptop GPU, and 32~GB RAM, running Ubuntu 22.04.

\subsection{2D Image Detection and Segmentation Evaluation}
\textbf{Detection Model Training} We fine-tuned the YOLO-World model for two different segmentation steps: the instance-level model was trained on 200 multi-view images with bounding box annotations, covering two categories, TRANSFER SYSTEM and ROBOT ARMS, for 75 epochs, taking a total of 15 minutes. The part-level model was trained on 600 multi-view images across six categories: RA (robot arm body), DRESSER, BOX, GUN, STAND, and BASE, also for 75 epochs, with a total training time of about 20 minutes.

\textbf{Detection and Segmentation Evaluation} The proposed framework leverages 2D masks to guide 3D point cloud segmentation, making 2D accuracy critical to overall performance. We evaluate two YOLO-World models and assess their integration with SAM. For comparison, we include a baseline that performs part-level segmentation directly using a single YOLO-World model with SAM, without instance-level guidance. As shown in Table~\ref{tab:2Dmask_mAP}, our hierarchical method performs higher part-level mask accuracy and better handles small structures and occlusions.

\begin{table}[!htbp]
\centering
\label{tab:2Dmask_mAP}

\resizebox{\textwidth}{!}{%
\begin{tabular}{ll|ccccccc|ccccccc}
\toprule
\multicolumn{2}{c|}{\textbf{Method}} 
& \multicolumn{7}{c|}{\textbf{mAP@0.5}} 
& \multicolumn{7}{c}{\textbf{mAP@0.5:0.95}} \\
\cmidrule(lr){3-9} \cmidrule(lr){10-16}
\textbf{Model} & \textbf{Strategy} 
& RA & Dresser & Box & Stand & Base & GUN & Transfer 
& RA & Dresser & Box & Stand & Base & GUN & Transfer \\
\midrule
YOLO-World      & Hierarchical & 0.99 & 0.96 & 0.99 & 0.98 & 0.97 & 0.95 & 0.90 & 0.86 & 0.81 & 0.88 & 0.77 & 0.73 & 0.76 & 0.75 \\
YOLO-World      & Single       & 0.90 & 0.70 & 0.71 & 0.60 & 0.58 & 0.52 & 0.75 & 0.80 & 0.59 & 0.60 & 0.53 & 0.51 & 0.39 & 0.59 \\
YOLO-World+SAM  & Hierarchical & 0.99 & 0.71 & 0.75 & 0.89 & 0.69 & 0.77 & 0.80 & 0.99 & 0.23 & 0.89 & 0.26 & 0.26 & 0.42 & 0.50 \\
YOLO-World+SAM  & Single       & 0.87 & 0.60 & 0.61 & 0.55 & 0.49 & 0.46 & 0.69 & 0.74 & 0.14 & 0.28 & 0.15 & 0.12 & 0.20 & 0.46 \\
\bottomrule
\end{tabular}
}  

\vspace{0.5em}
\caption{Comparison of mAP@0.5 and mAP@0.5:0.95 for each model and strategy across object categories.}
\label{tab:2Dmask_mAP}
\end{table}

\begin{figure}[!htbp]
  \centering
  \includegraphics[width=0.95\linewidth]{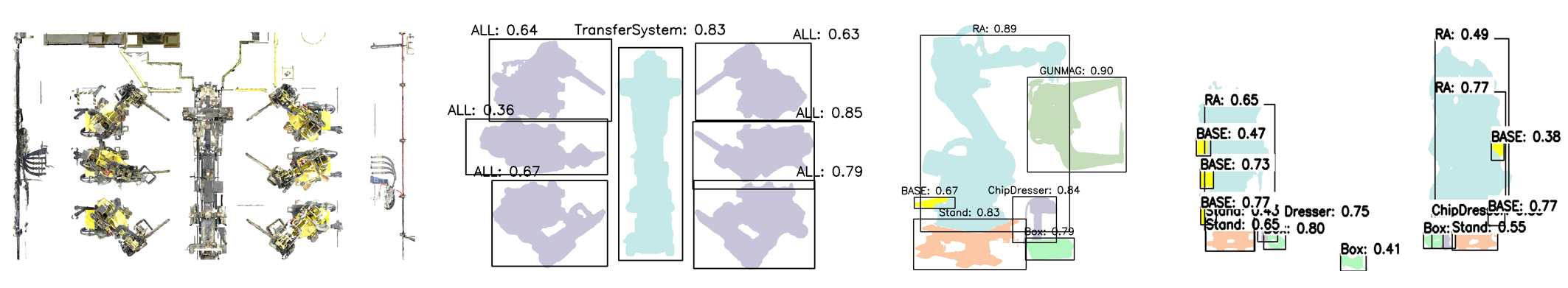}
  \vspace{0.5em}
  \caption{2D segmentation comparison, From left to right: (a) Input image (b) Instance segmentation (c) Part-level segmentation (d) Single stage segmentation}
  \label{fig:topview_comparison}
\end{figure}

\begin{figure}[!htbp]
  \centering
  \includegraphics[width=0.95\linewidth]{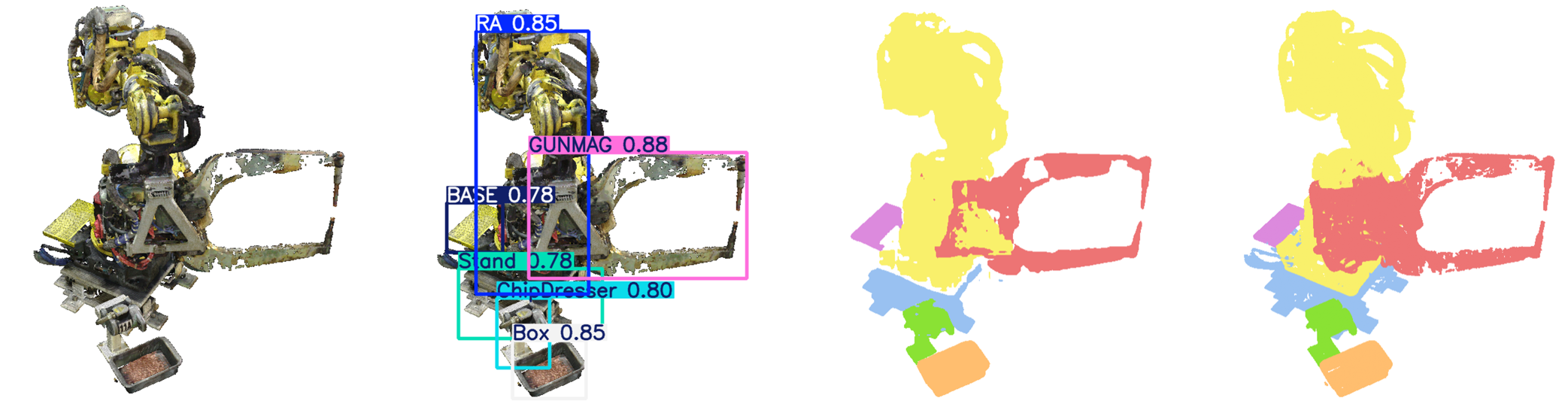}
  \vspace{0.5em}
  \caption{Fine-grained part segmentation results. From left to right: (a) Rendered image (b) YOLO-World Detection (c) YOLO-World+SAM segmentation (d) GT.}
  \label{fig:qualitative_comparison_transposed}
\end{figure}

\textbf{Hierarchical Segmentation} Figure~\ref{fig:topview_comparison} shows the 2D results of the hierarchical segmentation for the factory scene and the comparison with the single-step segmentation. It can be seen that the hierarchical segmentation has better segmentation results in both top-down and surround view, but the single model has obvious segmentation errors when segmenting part level objects in full view. In addition, we show the steps of part-level detection-by-segmentation in Figure~\ref{fig:qualitative_comparison_transposed} compared with ground truth. Additional multi-view results are provided in the appendix.

\subsection{3D Segmentation Results and Evaluation}
\begin{figure}[!htbp]
  \centering
  \includegraphics[width=0.2\linewidth]{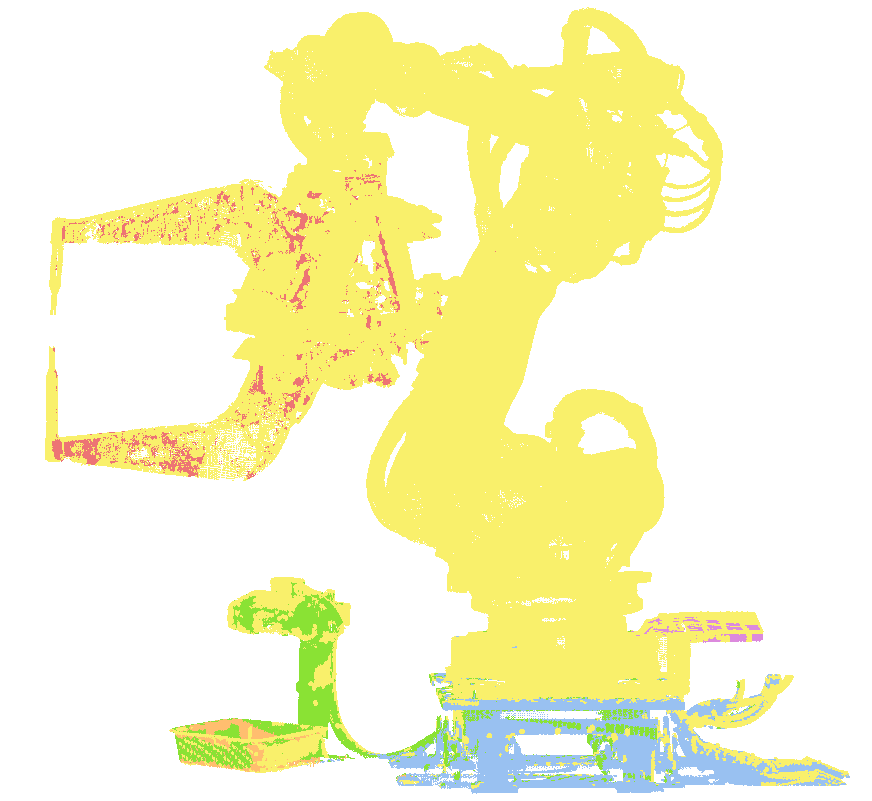}
  \hspace{1em}
  \includegraphics[width=0.2\linewidth]{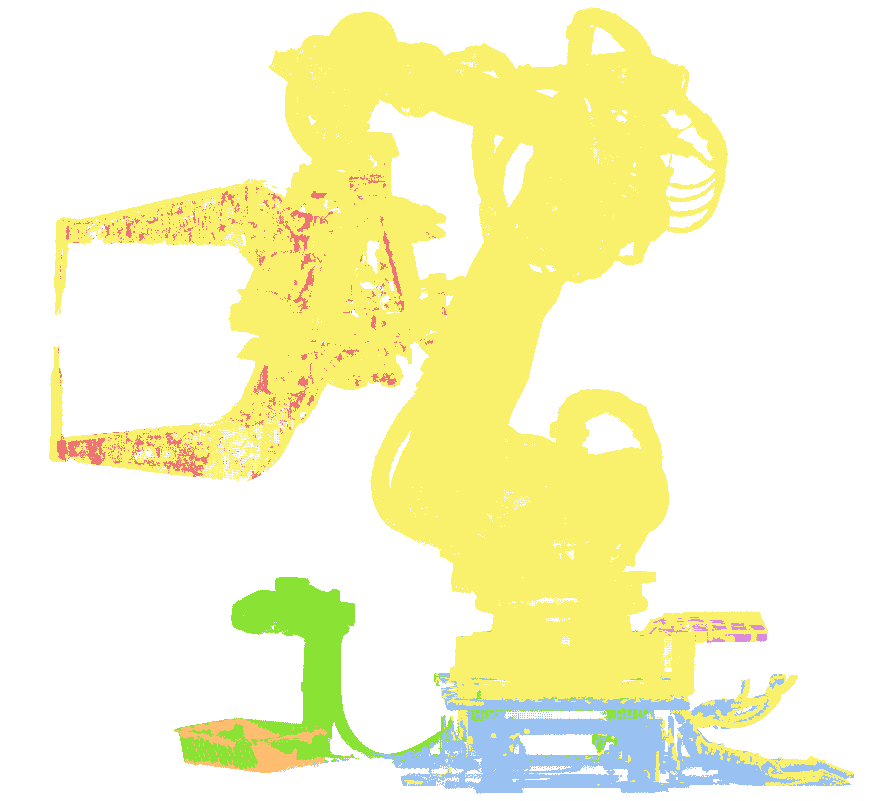}
  \hspace{1em}  
  \includegraphics[width=0.2\linewidth]{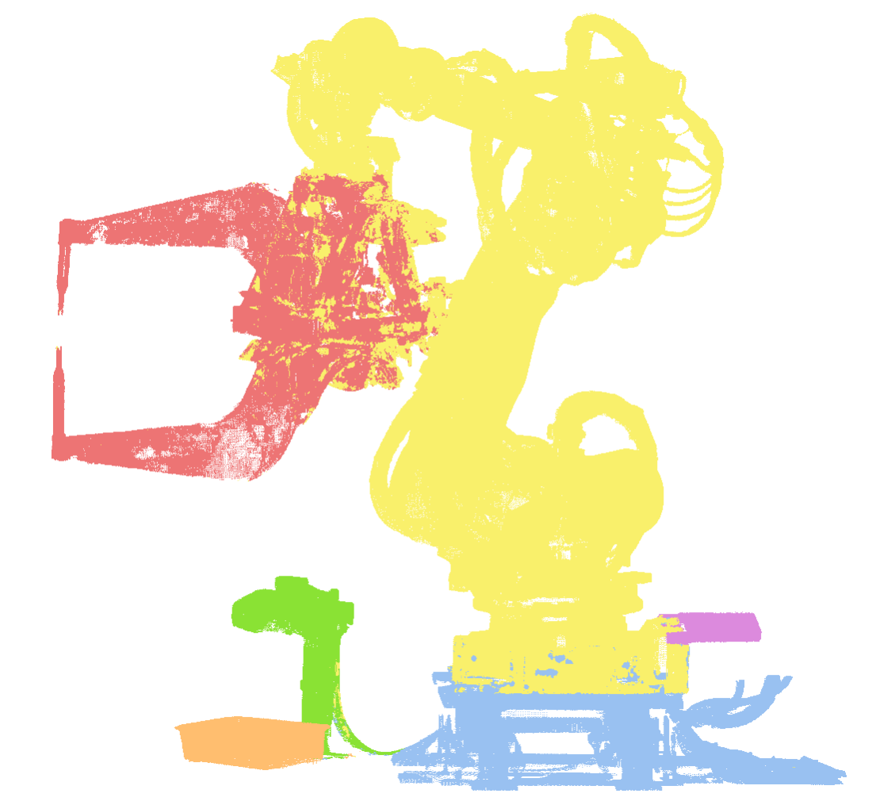}
  \hspace{1em}
  \includegraphics[width=0.2\linewidth]
  {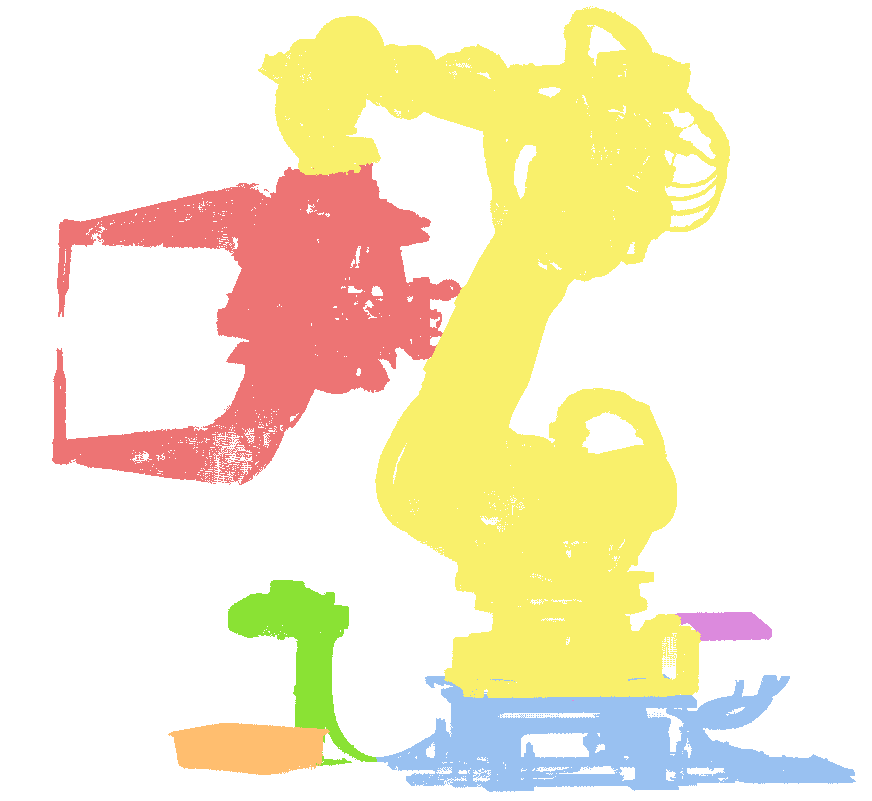}
  \label{(a) }
  \hspace{1em}  
  \vspace{0.5em}
  \caption{3D point cloud segmentation results. From left to right: (a) Projection-only (b) With cluster method (c) With Bayes (d) GT}
  \label{fig:qualitative_comparison_3d}
\end{figure}

We validate the effectiveness of our proposed Bayesian fusion method in addressing feature inconsistency and occlusion caused by multi-view projections through ablation studies. Due to the complex structure of the robotic arm, we sample 10 surrounding views and one top-down view. As shown in Figure~\ref{fig:qualitative_comparison_3d}(a), directly projecting multi-view 2D masks onto the 3D point cloud results in incorrect segmentation, as the robotic arm occludes the chipdresser or GUN in some views, and the chipdresser occludes the box. Figure~\ref{fig:qualitative_comparison_3d}(b) shows results with clustering refinement, which removes some noise but still struggles with occlusion, especially for the GUN. As shown in Figure~\ref{fig:qualitative_comparison_3d}(c), the Bayesian fusion result has clearer boundaries. Compared with the ground truth in Figure~\ref{fig:qualitative_comparison_3d}(d), the robot arm and GUN are segmented more accurately, and the box is no longer confused with the chipdresser. 

The evaluation metrics in Table~\ref{tab:3d_comparsion} further confirm that Bayesian fusion effectively handles occlusion and unclear boundaries, enabling accurate part-level segmentation. However, since there is no clear division between RA, GUN, and STAND at the connection position, the segmentation accuracy of these three categories is not particularly ideal. More visual results are provided in the appendix.



\begin{table}[!htbp]
\centering
\label{tab:3d_comparsion}

\resizebox{\textwidth}{!}{%
\begin{tabular}{l|ccc|ccc|ccc}
\hline
\toprule
& \multicolumn{3}{c|}{\textbf{Precision}} & \multicolumn{3}{c|}{\textbf{Recall}} & \multicolumn{3}{c}{\textbf{mIoU}} \\
\textbf{Objects} & Bayes & Cluster & Original & Bayes & Cluster & Original & Bayes & Cluster & Original  \\
\midrule
RA     & \textbf{67.42} &  56.31  & 54.67 & \textbf{98.03}         & 94.13     & 94.25   & \textbf{65.29} & 53.84  & 52.67 \\
Dresser & \textbf{98.96}         &{86.31} & 67.94  &\textbf{97.28}         & 88.24         &{88.97}  & \textbf{96.07} & 73.91  & 58.73 \\
Box     & \textbf{99.03}        & {97.84} & 89.26 & \textbf{99.12}         & 78.52         & 78.84   & \textbf{98.01} & 75.43 & 68.06 \\
Stand   & \textbf{87.69}         & 78.50  & 71.36 & \textbf{97.23}        & {86.20}        &{86.37}  & \textbf{83.43} & 68.06  & 63.93 \\
Base    & \textbf{96.24} &         86.83  & 83.92 & \textbf{99.31}         & 75.84         & 75.29   & \textbf{96.01} & 66.67  & 63.93 \\
GUN     & \textbf{98.24} &         79.66  & 59.32          &67.89  &\textbf{93.00} &{92.14}   &{57.48} &\textbf{ 72.43} & 55.03 \\
\bottomrule
\end{tabular}
}
\vspace{0.5em}
\caption{Per-category comparison of Precision(\%), Recall(\%), and mIoU(\%) for each segmentation strategy (Bayes, Cluster, and Original).}
\label{tab:3d_comparsion}
\end{table}

\subsection{Final Segmentation Results in Full Scene}

\begin{figure}[!htbp]
  \centering
  \includegraphics[width=0.85\linewidth]{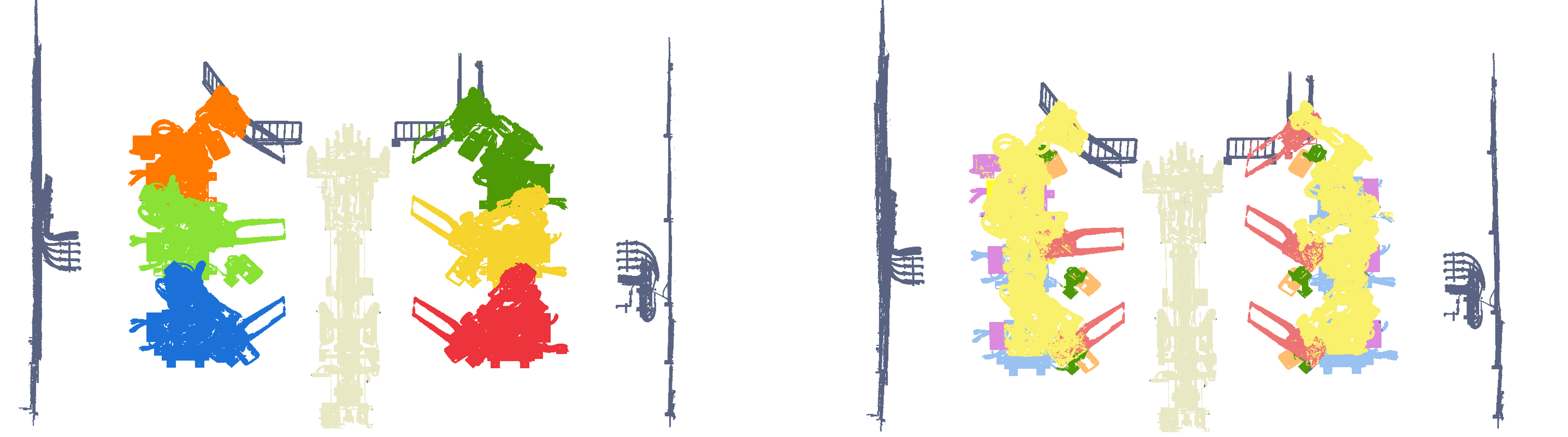}
   \vspace{0.4em}
  \caption{Final 3D segmentation results. (Left) Instance-level segmentation with each robot arm and transfer system shown in a distinct color. (Right) Part-level segmentation after Bayesian fusion, with fine-grained components accurately labeled.}
  \label{fig:results}
\end{figure}

Figure~\ref{fig:results} illustrates the complete segmentation outcome across the full industrial environment. The instance-level result clearly separates robot arms and transfer systems, enabling structure-aware spatial partitioning. The final part-level segmentation  reflects high-quality recognition of individual components across all stations. Our framework maintains consistent part labeling despite complex scenes and varying viewpoints, demonstrating that our framework achieves accurate 3D segmentation in industrial scenes.

\subsection{Generalization Evaluation on Public Datasets}
\begin{figure}[!htbp]
  \centering
  \includegraphics[width=0.8\linewidth]{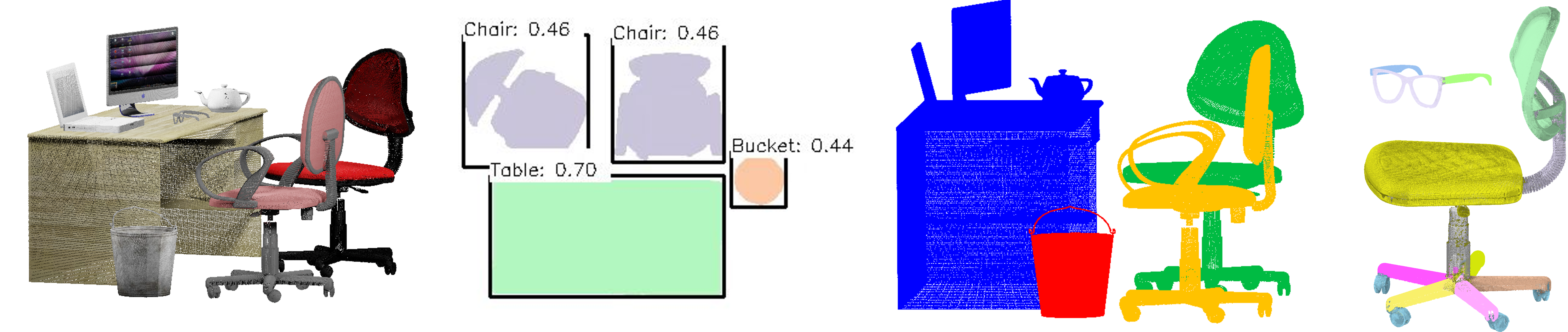}
  \vspace{0.4em}
  \caption{3D segmentation based on PartNet dataset. From left to right: (a) Composite scene (b) 2D Instance masks (c) 3D Instances (d) 3D part segmentation results.}
  \label{fig:partnet_scene1}
\end{figure}

To evaluate the generalization ability of our framework in common 3D scenes, we conduct experiments on the PartNet dataset~\cite{DATASET_partnet}. Since each object in PartNet is provided as an individual instance, we create multi-object scenes by combining several objects to increase scene complexity, as shown in Figure~\ref{fig:partnet_scene1}. We apply our hierarchical framework for part-level segmentation to the composite scene, demonstrating its effectiveness in general environments. 

\begin{figure}[!htbp]
  \centering
  \includegraphics[width=0.9\linewidth]{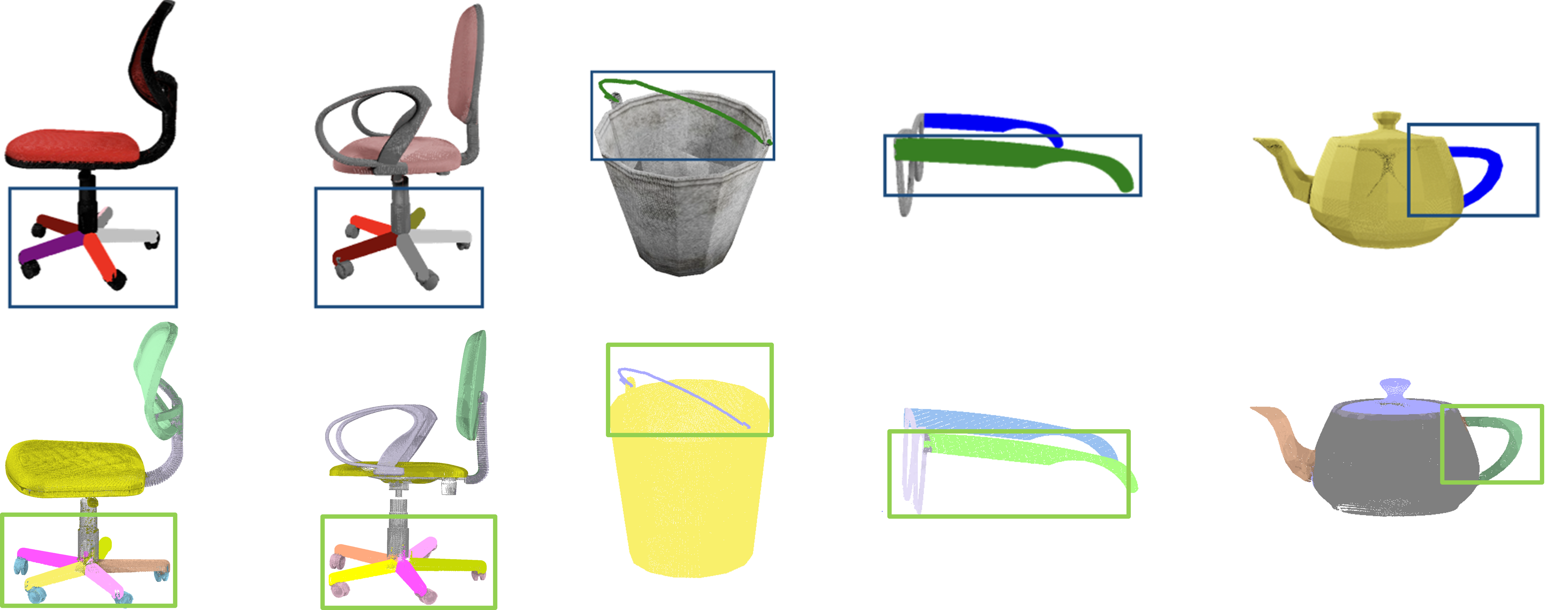}
  \vspace{0.4em}
  \caption{Comparison of results with part segmentation of PartSLIP++. Row (1) illustrate the results from PartSLIP++, and Row (2) from our method.}
  \label{fig:comparsion_partnet}
\end{figure}

Since the PartNet~\cite{DATASET_partnet} dataset consists of common object categories, we directly use GLIP~\cite{GLIPv2} as the detection model without fine-tuning on a custom dataset. Figure~\ref{fig:comparsion_partnet} compares our results with those of PartSLIP++~\cite{partslip++}, where different colors indicate different parts. Notably, we use the same GLIP+SAM pipeline as PartSLIP++ for 2D mask generation. As shown in Table~\ref{tab:category_iou}, our part segmentation in full scenes achieves accuracy close to that of other methods segmenting parts from individual objects. These results further validate the effectiveness of our Bayesian fusion and demonstrate the generalization and adaptability of our framework. For more experimental results, please refer to the appendix.

\begin{table}[!htbp]
\centering
\begin{tabular}{lcccc}
\toprule
\textbf{Method} & \textbf{Chair} & \textbf{Kettle} & \textbf{Eyeglasses} & \textbf{Bucket} \\
\midrule
PointNet++\cite{3D_POINTNET++}    & 42.20 & 20.90 & 78.65   & 0.00 \\
PartSLIP\cite{partslip}      & \textbf{85.30} & 77.00 & 77.00   & \textbf{89.6} \\
PartSLIP++\cite{partslip++}    & 85.30 & \textbf{85.60} & 88.25   & 85.50 \\
Ours                         & 78.71 & 75.96 & \textbf{89.38}   & 86.46 \\
\bottomrule
\end{tabular}
\vspace{0.5em}
\caption{Per-category segmentation mIoU (\%) comparison on segmented objects.}
\label{tab:category_iou}
\end{table}

\section{Conclusion and Future Works}
We present a 2D detection-by-segmentation-guided hierarchical 3D segmentation framework. The detection results generated by YOLO-World provide labels and prompts for SAM, achieving accurate segmentation across both object and part levels. We also utilize the Bayesian fusion method to resolve cross-view inconsistencies and optimize 3D segmentation results. Experiments on real-world factory data and general open source dataset confirm robust segmentation under occlusion and clutter, achieving high precision across categories.

However, it is still challenging to deal with small or severely occluded parts, especially objects with unclear boundary features, which are difficult to accurately segment. To solve these problems, future work will consider incorporating information such as depth into the segmentation model to provide richer feature expressions. In addition, multi-view consistency will be attempted to be modeled into the recognition and segmentation models to ensure the consistency of multiple views at the 2D segmentation level, thereby avoiding additional errors introduced in the subsequent fusion projection step.
\par\bigskip  
\noindent
\textbf{Acknowledgment.} This work was supported by JST SPRING, Grant Number JPMJSP2114.

\end{document}